\documentclass[preprint,11pt]{elsarticle}

\usepackage{amssymb}

\journal{ }
\usepackage{graphicx}
\usepackage{amsmath}
\usepackage{multirow}
\usepackage{url}
\usepackage{subfig}
\usepackage{multicol}
\usepackage{geometry}
\begin{document}

\begin{frontmatter}



\title{\textbf{BP-STDP: Approximating Backpropagation using\\ Spike Timing Dependent Plasticity}}


\author{Amirhossein Tavanaei, Anthony Maida}

\address{The Center for Advanced Computer Studies\\University of Louisiana at Lafayette\\
              Louisiana, Lafayette, LA 70504, USA\\
              \{tavanaei,maida\}@louisiana.edu}

\begin{abstract}
The problem of training spiking neural networks (SNNs) is a necessary precondition to
understanding computations within the brain, a field still in its infancy.
Previous work has shown that supervised learning in multi-layer SNNs enables 
bio-inspired networks to recognize patterns of stimuli through hierarchical feature
acquisition. Although gradient descent has shown impressive performance in
multi-layer (and deep) SNNs, it is generally not considered biologically plausible
and is also computationally expensive. This paper proposes a novel supervised learning approach based on an event-based spike-timing-dependent plasticity (STDP) rule embedded in a network of integrate-and-fire (IF) neurons. The proposed temporally local learning rule follows the backpropagation weight change updates applied at each time step. This approach enjoys benefits of both accurate gradient descent and temporally local, efficient STDP. Thus, this method is able to address some open questions regarding accurate and efficient computations that occur in the brain. The experimental results on the XOR problem, the Iris data, and the MNIST dataset demonstrate that the proposed SNN performs as successfully as the traditional NNs. Our approach also compares favorably with the state-of-the-art multi-layer SNNs.
\end{abstract}

\begin{keyword}
Spiking neural networks, STDP, Supervised learning, Temporally local learning, Multi-layer SNN.

\end{keyword}

\end{frontmatter}


\section{Introduction}
In the vein of neural network research, spiking neural networks (SNNs) have attracted recent and long-standing interest due to their biologically plausible realism, theoretical computation power, and power efficiency~\cite{maass1997networks,ghosh2009spiking,maass2015spike,neil2016learning,kasabov2013dynamic}.
Neurons in SNNs (analogous to biological neurons in the brain) communicate via discrete spike events. An important question in implementing spiking frameworks is how these networks are trained under supervision while there is no differentiable activation function of continuous values? Multi-layer supervised learning is a crucial concept in SNNs to prove their ability to predict and classify the patterns of stimuli. 
 
The earliest supervised learning in SNNs, SpikeProp, was proposed by Bohte et al. (2002)~\cite{bohte2002error} which develops a gradient descent based learning method similar to traditional backpropagation embedded in multi layer neural networks. SpikeProp minimizes the distance between single desired and output spikes by formulating the spike time as a function of the neuron's membrane potential. Later, Quick-Prop and RProp~\cite{mckennoch2006fast} were introduced as faster versions of SpikeProp. Following the same approach, Booij et al. (2005)~\cite{booij2005gradient} and Ghosh et al. (2009)~\cite{ghosh2009new} proposed new SpikeProp algorithms minimizing the distance between multiple desired and output spikes to improve the model's coding ability in response to temporal patterns. Gradient descent (GD), as a popular supervised learning approach, is still being employed to develop high performance supervised learning in spiking platforms either in offline~\cite{xu2013supervised} or online~\cite{xu2017online} manner. Chronotron~\cite{florian2012chronotron} which utilizes the Victor \& Purpura (VP) metric~\cite{victor1997metric} for E-learning (offline) and the synaptic current simulation for I-learning (online) is another example of GD in SNNs. As implemented in Chronotron, the error function in GD can be defined by a difference measure between the post- and presynaptic current (or membrane potential)~\cite{huh2017gradient} (online) or a spike train distance metric (offline) such as von Rossum distance~\cite{van2001novel,zenke2017superspike} and inner product of spike trains~\cite{lin2017supervised}. The online gradient descent method has also attracted recent interest in deep SNNs~\cite{lee2016training,wu2017spatio}. Although the online GD methods have been successful in developing supervised learning in multi-layer SNNs, using membrane potential and derivative based approaches are biologically implausible because SNNs only communicate via discrete spike events. Additionally, GD is computationally expensive for multi-layer SNNs. Recently, Xie et al. (2016)~\cite{xie2016efficient} developed a normalized spiking backpropagation calculating postsynaptic spike times (which still needs expensive computations) instead of membrane potential at each time step to improve the algorithm's efficiency.        

Another vein of research utilizes modified versions of the Widrow-Hoff learning rule for SNNs. For instance, ReSuMe~\cite{ponulak2010supervised} introduced a single-layer supervised learning approach; and later, it was extended to the multi-layer framework using backpropagation by Sporea et al. (2013)~\cite{sporea2013supervised}. SPAN~\cite{mohemmed2012span,mohemmed2013training} also used Widrow-Hoff by transforming the discrete spikes to corresponding analog signals. The learning methods mentioned above are more efficient but less accurate than the GD-based approaches. Another efficient, supervised learning method belongs to the perceptron-based approaches such as Tempotron~\cite{gutig2006tempotron} where each spike event is treated as a binary tag for training the perceptron~\cite{yu2014brain,xu2013new}. These models present single layer supervised learning. However, the idea of spike/no-spike classification broadens a new supervised learning category incorporating efficient spike-timing-dependent plasticity (STDP)~\cite{markram2012spike,song2000competitive,caporale2008spike} and anti-STDP that are triggered according to the neuron's label~\cite{wang2014online,tavanaei2017spiking}. STDP is a biologically plausible learning rule occurs in the brain in which the presynaptic spikes occur immediately before the current postsynaptic spike strengthen the interconnecting synapses (LTP); otherwise, the synapses are weakened (LTD). 

In this paper, we propose novel multi-layer, supervised learning rules to train SNNs of integrate-and-fire (IF) neurons. The proposed approach takes advantages of the both efficient, bio-inspired STDP and high performance backpropagation (gradient descent) rules. First, we show that the IF neurons approximate the rectified linear units. Then, we develop a temporally local learning approach specified by an STDP/anti-STDP rule derived from the backpropagation weight change rules that can be applied at each time step. 

\section{Method}
Before proposing the spike-based, temporally local learning rules, we show how the biological IF neurons approximate the well-known artificial neurons equipped with the rectified linear unit activation function. This approximation takes the first step in converting rate-based learning to spatio-temporally local learning. 

\subsection{Rectified Linear Unit versus IF neuron}
A neuron with the rectified linear unit (ReLU) activation function, $f(y)$, receiving input signals, $x_h$, via corresponding synaptic weights, $w_h$, is defined as follows
\begin{equation}
f(y) = \max (0,y) \ , \ \ y=\sum_h x_h w_h  
\label{eq:relu}
\end{equation}
\begin{equation}
\frac{\partial f}{\partial y} = 
\begin{cases}
1, & y>0 \\
0, & y\leq 0
\end{cases}
\label{eq:reluderivation}
\end{equation}

\noindent
\textbf{Theorem: }The IF neuron approximates the ReLU neuron. Specifically, the membrane potential of the IF neuron approximates the activation value of the ReLU neuron. 

\noindent
\textbf{Proof: }A non-leaky IF neuron integrates the temporal membrane potentials caused by its input spike trains and fires when its membrane potential, $U(t)$, reaches the neuron's threshold, $\theta$. A simplified formulation of the IF neuron is shown in Eq.~\ref{eq:if}. 
\begin{subequations}
\label{eq:if}
\begin{align}
&U(t) = U(t-\Delta t) + \sum_h w_h(t) s_h(t)\\
&\mathrm{if} \ \ U(t) \geq \theta  \ \ \mathrm{then} \ \ r(t)=1, \ U(t)=U_{\mathrm{rest}} 
\end{align}
\end{subequations}
Where, $s_h(t)$ and $r(t)$ are pre- and postsynaptic spikes at time $t$ ($s_h(t),r(t) \in \{ 0,1\}$), respectively. The neuron's membrane potential is reset to resting potential, $U_{\mathrm{rest}}$, (assume $U_{\mathrm{rest}}=0$) upon firing. By formulating the presynaptic spike train, $G_h(t)$, as the sum of delta Dirac functions (Eq.~\ref{eq:prespike}), the input value, $x_h\in [0,1]$, in Eq.~\ref{eq:relu}, can be determined by integrating over the spiking time interval, $T$, as shown in Eq.~\ref{eq:spikepreintgral}.
\begin{equation}
\label{eq:prespike}
G_h(t) = \sum_{t_h^p \in \{ s_h(t)=1\} } \delta (t-t_h^p)
\end{equation}
\begin{equation}
\label{eq:spikepreintgral}
x_h = \frac{1}{K} \int_0^T G_h(t^{\prime}) dt^{\prime}
\end{equation}
$K$ is a constant denoting the maximum number of spikes in $T$ ms interval. 

\begin{figure}
\centering
\includegraphics[scale=.55]{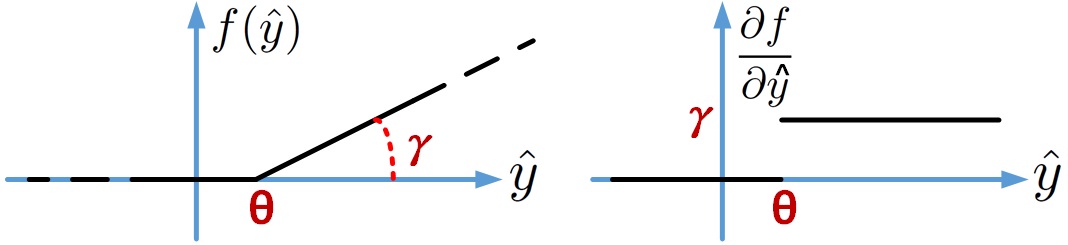}
\caption{IF activation function ($f(\hat{y})$) and its derivative. This function resembles a scaled version of the ReLU function that is shifted by $\theta$. }
\label{fig:activefunc}
\end{figure}

In a short time period $(t-\alpha,t]$ between consecutive postsynaptic spikes $r(t-\alpha)$ and $r(t)$, the neuron's membrane potential, following Eq.~\ref{eq:if}, is obtained by
\begin{equation}
\label{u_1}
U(t) = \sum_h w_h \big ( \int_{t-\alpha^+}^t \sum_{t_h^p} \delta(t^{\prime}-t_h^p)dt^{\prime} \big )
\end{equation} 
The membrane potential calculated above ($U(t)$) is greater than the threshold, $\theta$ ($U(t)\geq \theta$). Thus, as we assumed that the IF neuron is non-leaky and its membrane potential resets to zero upon firing, $R$ postsynaptic spikes demand an accumulated membrane potential (Eq.~\ref{eq:u_2}). The accumulated membrane potential, $U^{\mathrm{tot}}$, is obtained by linear summation over the sub-membrane potentials computed in Eq.~\ref{u_1}. By this definition, the postsynaptic spike count, $R$, represents the output value, $\hat{y}$, that is passed through a non-linear, threshold-based activation function.  
\begin{equation}
\label{eq:u_2}
U^{\mathrm{tot}} = \hat{y} = \sum_{t^{\mathrm{f}}\in \{ r(t)=1\} } U(t^{\mathrm{f}})
\end{equation}
$U^{\mathrm{tot}}$ specifies the IF neuron's activity in $T$ ms, which is proportionally related to the postsynaptic spike count, $R$. Therefore, the activation function of the IF neuron can be expressed as
\begin{equation}
\label{activeif}
f(\hat{y}) = 
\begin{cases}
R=\gamma \hat{y} & \hat{y}>\theta \\
0 & \mathrm{otherwise}
\end{cases}
\end{equation}
where, $\gamma \propto T\cdot \theta^{-1}$ is a constant controlling the postsynaptic spike count. This activation function is similar to a linearly scaled ReLU function that is shifted to right by $\theta$. Figure~\ref{fig:activefunc} shows the activation function and its derivative.

This theorem opens a new door to develop new spike-based learning rules (applied to the spiking IF neurons) derived from the traditional learning rules (applied to the ReLU neurons). Specifically, in the next section, we use this theorem to propose an STDP-based backpropagation rule applied to the spiking IF neurons. 

\subsection{Backpropagation using STDP}
\begin{figure}
\centering
\includegraphics[scale=.32]{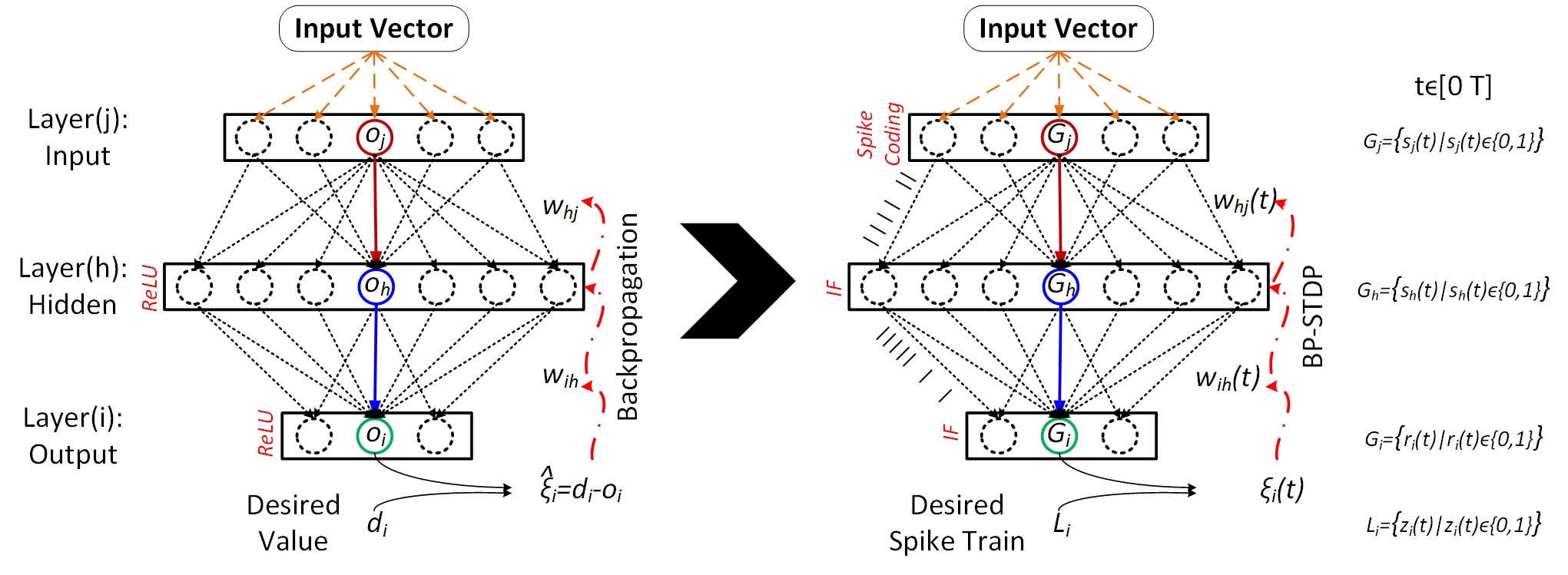}
\caption{The 2-Layer, conventional (left) and spiking (right) network architectures. The SNN receives spike trains representing input feature values in $T$ ms. The learning rules and the network status in the SNN are specified by an additional term as time ($t$). The formulas and parameters are discussed in Eqs.~\ref{eq:sgd} through~\ref{weightUpdate}.}
\label{fig:nets}
\end{figure}

The proposed learning rules are inspired from the backpropagation update rules reported for neural networks that are equipped with ReLU activation function. Figure~\ref{fig:nets} shows the network architectures and parameters used to describe the conventional and spiking neural networks in this paper. The main difference between these two networks is their data communication where the neural network (left) receives and generates real numbers while the SNN (right) receives and generates spike trains in $T$ ms time intervals.

Non-spiking neural networks equipped with gradient descent (GD) solve an optimization problem in which the squared difference between the desired, $d$, and output, $o$, values is minimized~\cite{bishop2006pattern,goodfellow2016deep}. A common objective function computed for $M$ output neurons receiving $N$ training samples is shown in Eq.~\ref{eq:sgd}.
\begin{equation}
\label{eq:sgd}
E = \frac{1}{N} \sum_{k=1}^N \sum_{i=1}^M (d_{k,i}-o_{k,i})^2
\end{equation}   
The weight change formula (using GD with learning rate of $\mu$) for a linear output neuron, say $i$, receiving $H$ inputs, $o_h$, (for a single training sample) is achieved by
\begin{equation}
E=(d_i-o_i)^2 = \big ( d_i-\sum_h o_h w_{ih} \big )^2 \rightarrow \ 
\frac{\partial E}{\partial w_{ih}} = -2 (d_i-o_i)\cdot o_h
\end{equation}
By reversing the sign on the derivative, we have
\begin{equation}
\label{outsgd}
\Delta w_{ih} \propto -\frac{\partial E}{\partial w_{ih}} \  \rightarrow \ \Delta w_{ih} = \mu (d_i-o_i)o_h
\end{equation}
By assuming $d_i,o_i,$ and $o_h$ as the spike counts of spike trains $L_i,G_i,$ and $G_h$~\cite{tavanaei2017representation} (see Eqs.~\ref{eq:prespike} and~\ref{eq:sgdspiketrain}), respectively, the weight change, defined above, can be re-expressed such that it computes the synaptic weight update in an SNN. This assumption is valid upon approximating the spiking IF neurons to the ReLU neurons (Theorem 1). Eq.~\ref{eq:sgdsnn} shows this update rule after $T=50$ ms.
\begin{subequations}
\begin{align}
&G_i(t) = \sum_{t_i^p \in \{ r_i(t)=1\} } \delta (t-t_i^p)\\
&L_i(t) = \sum_{t_i^q \in \{ z_i(t)=1\} } \delta (t-t_i^q)
\end{align}
\label{eq:sgdspiketrain}
\end{subequations}
\begin{equation}
\label{eq:sgdsnn}
\begin{aligned}
\Delta w_{ih} = \mu \int_0^T \big ( L_i(t^{\prime}) - G_i(t^{\prime})   \big )dt^{\prime} \cdot 
\int_0^T G_h(t^{\prime}) dt^{\prime}
\end{aligned}
\end{equation}

However, the weight change rule in Eq.~\ref{eq:sgdsnn} is not local in time. To make the learning rule local in time, we break the time interval, $T$, into sub-intervals such that each sub-interval contains zero or one spike. Hence, the learning rule, in a short time period of Eq.~\ref{eq:sgdsnn}, is specified by Eq.~\ref{eq:sgd1}.
\begin{equation}
\label{eq:sgd1}
\Delta w_{ih}(t) \propto \mu \big (z_i(t) - r_i(t)\big ) s_h(t)
\end{equation}
To implement the formula above, a combination of event-based STDP and anti-STDP can be used. The proposed learning rule updates the synaptic weights using a teacher signal to switch between STDP and anti-STDP. That is, the target neuron undergoes STDP and the non-target ones undergo anti-STDP. The desired spike trains, $\mathbf{z}$, are defined based on the input's label. Therefore, the target neuron is represented by a spike train with maximum spike frequency ($\beta$) and the non-target neurons are silent. Additionally, the learning rule triggers at desired spike times, $z_i(t)$ (the desired spike times are the same for all the target neurons). Eq.~\ref{eq:ruleoutput} shows the weight change that is applied to the output layer of our supervised SNN.
\begin{equation}
\label{eq:ruleoutput}
\Delta w_{ih} (t) = \mu \cdot \xi_i(t) \sum_{t^{\prime} = t-\epsilon}^t s_h(t^{\prime}) 
\end{equation}
\begin{equation}
\label{eq:zeta}
\xi_i(t) = 
\begin{cases}
1, & z_i(t) = 1, \ r_i\neq1 \ \mathrm{in} \  [t-\epsilon, t] \\
-1, & z_i(t) = 0, \ r_i=1 \ \mathrm{in} \  [t-\epsilon, t]\\
0, & \mathrm{otherwise}
\end{cases}
\end{equation}
Then, the synaptic weights of output layer are updated by
\begin{equation}
\label{eq:updateW1}
w_{ih}(t) = w_{ih}(t) + \Delta w_{ih}(t)
\end{equation}
The target neuron is determined by $z_i(t)$ where $z_i(t) = 1$ denotes the target neuron and $z_i(t) = 0$ denotes the non-target neuron.
The weight change scenario for the output layer starts with the desired spike times. At desired spike time $t$, the target neuron should fire in a short time interval $[t-\epsilon,t]$ known as the STDP window. Otherwise, the synaptic weights are increased proportionally by the presynaptic activity (mostly zero or one spike) in the same time interval. The presynaptic activity is denoted by $\sum_{t^{\prime} = t-\epsilon}^t s_h(t^{\prime})$ that counts the presynaptic spikes in $[t-\epsilon, t]$ interval. On the other hand, the non-target neurons upon firing undergo weight depression in the same way. This scenario is inspired from the traditional GD while supporting spatio-temporal, local learning in SNNs.

The learning rule written above works for a single layer SNN trained by supervision. To train a multi-layer SNN, we use the same idea that is inspired from the traditional backpropagation rules. The backpropagation weight change rule applied to a hidden layer of ReLU neurons is shown in Eq.~\ref{eq:bphidden}. 
\begin{equation}
\label{eq:bphidden}
\Delta w_{hj} = \mu \cdot \big (\sum_i \hat{\xi}_{i} w_{ih} \big ) \cdot o_j \cdot [o_h>0]
\end{equation}   
Where, $\hat{\xi}_{i}$ denotes the difference between the desired and output values ($d_i-o_i$). In our SNN, $\hat{\xi}_{i}$ is approximated by $\xi_i$ (Eq.~\ref{eq:zeta}). The value $[o_h>0]$ specifies the derivative of ReLU neurons in the hidden layer. Using the approximation of the IF neurons to the ReLU neurons (Eq.~\ref{activeif}), similar to the output layer (Eq.~\ref{eq:sgdsnn}), the weight change formula can be re-expressed in terms of spike counts in a multi-layer SNN as shown in Eq.~\ref{eq:spikecounthidden}.
\begin{align}
\label{eq:spikecounthidden}
\Delta w_{hj} = \mu \int_0^T \big ( \sum_i \xi_{i}(t^{\prime}) w_{ih}(t^{\prime}) \big ) dt^{\prime} \cdot 
\int_0^T \big ( \sum_{t_j^p} \delta (t^{\prime}-t_j^p) \big ) dt^{\prime}  \cdot \\ 
 \bigg ( \big [ \int_0^T \sum_{t_h^p} \delta (t^{\prime}-t_h^p) dt^{\prime} \big ] >0 \bigg ) \nonumber
\end{align}
After dividing $T$ into short sub-intervals $[t-\epsilon, t]$, the temporally local rule for updating the hidden synaptic weights is formulated as follows.
\begin{equation}
\label{eq:hiddenweight}
\Delta w_{hj}(t) = 
\begin{cases}
\mu \cdot \sum_i \xi_{i}(t) w_{ih}(t) \cdot \sum_{t^{\prime} = t-\epsilon}^t s_j(t^{\prime})
& ,s_h=1 \ \mathrm{in} \ [t-\epsilon,t] \\
0 & ,\mathrm{otherwise}
\end{cases}
\end{equation} 
Finally, the synaptic weights of hidden layer are updated by
\begin{equation}
\label{weightUpdate}
w_{hj}(t) = w_{hj}(t) + \Delta w_{hj}(t)
\end{equation} 

\begin{figure}
\centering
\small
\framebox{%
\vbox{
\begin{tabbing}
\= xxxxxxx\= xxxx \= xxxx \= xxxx \= xxxx \= xxxx \= xxxx\kill
\> 1: for $t=\Delta t : T: \Delta t$ do \\
\> \> - - First Layer \\
\> 2: \> $U_1[:,t] = \mathbf{s}_j[:,t]*\mathbf{w}_{jh} + U[:,t-\Delta t]$ \\
\> 3: \> for $u$ in hidden neurons do \\
\> 4: \> \> if $U_1[u,t]>\theta_h$ then \\
\> 5: \> \> \> $\mathbf{s}_h[u,t]=1$ \\
\> 6: \> \> \> $U_1[u,t] = U_{\mathrm{rest}}$ \\
\> \> - - Second Layer \\
\> 7: \> $U_2[:,t] = \mathbf{s}_h[:,t]*\mathbf{w}_{hi} + U_2[:,t-\Delta t]$ \\
\> 8: \> for o in output neurons do \\
\> 9: \> \> if $U_2[o,t]>\theta_o$ then \\
\> 10: \> \> \> $\mathbf{r}_i[o,t]=1$ \\
\> 11: \> \> \> $U_2[o,t]=U_{\mathrm{rest}}$ \\
\> \> - - Weight Adaptation  \\
\> 12: \>  if $t \in \mathbf{z}$ then \ \ - - $\mathbf{z}=\{ t|z(t)=1\}$ \\
\> 13: \> \> if $r_i=0$ in $[t-\epsilon, t]$ for target neuron: \\
\> 14: \> \> \> $\xi_i[target]$= 1 \\
\> 15: \> \> if $r_i=1$ in $[t-\epsilon, t]$ for non-target neurons: \\
\> 16: \> \> \> $\xi_i[non~targets]$= [-1] \\
\> 17: \> \> $derivatives_h = \big [\mathbf{s}_h(t)=1$ in $[t-\epsilon, t]\big ]$ \\
\> 18: \> \> $\xi_h=\mathbf{w}_{hi}*\xi_i \cdot derivatives_h$ \\
\> 19: \> \> $\mathbf{w}_{hi}+=$ sum$(\mathbf{s}_h(t-\epsilon...t))*\xi_i \cdot \mu$ \\
\> 20: \> \> $\mathbf{w}_{jh}+=$ sum$(\mathbf{s}_j(t-\epsilon...t))*\xi_h \cdot \mu$
\end{tabbing} } }
\caption{The BP-STDP algorithm applied to a multi-layer SNN consisting of input, output and one hidden layer. `*' stands for matrix multiplication. `- -' starts a comment.}
\label{fig:alg}
\end{figure}  

The above learning rule can be non-zero when the hidden neuron $h$ fires (postsynaptic spike occurrence). Thus, the weights are updated according to the presynaptic ($s_j(t)$) and postsynaptic ($s_h(t)$) spike times, analogous to the standard STDP rule. Additionally, the derivative of ReLU ($o_h>0$) is analogous to the spike generation in the IF neurons (see the condition in Eq.~\ref{eq:hiddenweight}). Following this scenario for the spatio-temporally synaptic weight change rule, we can build a multi-layer SNN equipped with the STDP-based Backpropagation algorithm, named BP-STDP. Figure~\ref{fig:alg} demonstrates the BP-STDP algorithm applied to an SNN with one hidden layer.

\section{Results}
We ran three different experiments to evaluate the proposed model (BP-STDP) on the XOR problem, the iris dataset~\cite{fisher1936use}, and the MNIST dataset~\cite{lecun1998mnist}. The parameters used in the experimental setting are shown in Table~\ref{tb:params}. The synaptic weights for all the experiments are initialized by the Gaussian random numbers with zero mean and unit standard deviation.

\begin{table}[]
\centering
\caption{Model parameters.}
\label{tb:params}
\begin{tabular}{|l|l|l|l|}
\hline
\multicolumn{1}{|c|}{\textbf{Parameter}} & \multicolumn{1}{c|}{\textbf{Value}} & \multicolumn{1}{c|}{\textbf{Parameter}} & \multicolumn{1}{c|}{\textbf{Value}} \\ \hline
$\epsilon$                                      & 4 ms                           & $\mu$                                      & 0.0005                              \\ \hline
$\beta$                                    & 250 Hz & $\theta_o$                                  & $0.025\times H$                                                                 \\ \hline
$\theta_h$                                   & 0.9                                 & $H$                                       & \{10,...,1500\}                    
 \\ \hline
$\Delta t$                                   & 1 ms & $U_0$                                       & 0                    
 \\ \hline
\end{tabular}
\end{table}

\subsection{XOR problem}
The BP-STDP algorithm is evaluated by the XOR problem to show its ability to solve linear inseparability. The dataset contains four data points $\{(0.2,0.2), (0.2,1) , (1,0.2), (1,1)\}$ and corresponding labels $\{0,1,1,0\}$. We used 0.2 instead of 0 to activate the IF neurons (to release spikes). The network architecture consists of 2 input, 20 hidden, and 2 output IF neurons. The number of hidden neurons for this problem has little effect on the results. We will investigate the impact of the number of hidden neurons for the MNIST classification task. Each input neuron releases spike trains corresponding to the input values such that the value 1 is represented by a spike train with the maximum spike rate (250 Hz).

\begin{figure}
\centering
\includegraphics[scale=.5]{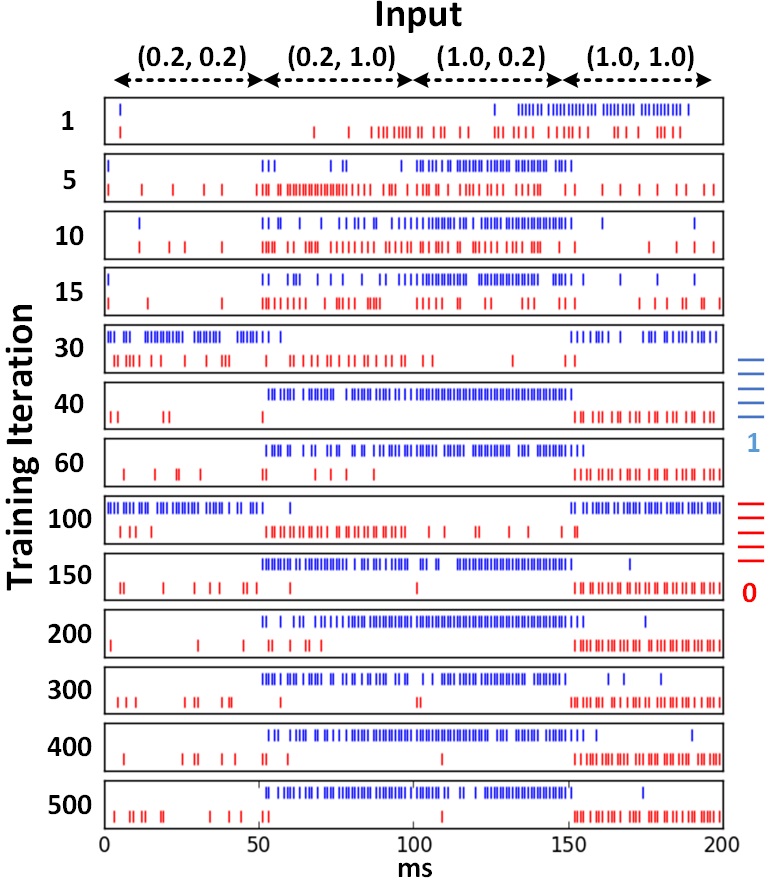}
\caption{Spike trains released from the output neurons in response to four pairs of spike trains representing $\{(0.2,0.2), (0.2,1) , (1,0.2), (1,1)\}$ values in 1 through 500 iterations. We used high spike rates for better visualization.}
\label{fig:xorspike}
\end{figure}

\begin{figure}
\centering
\includegraphics[scale=.44]{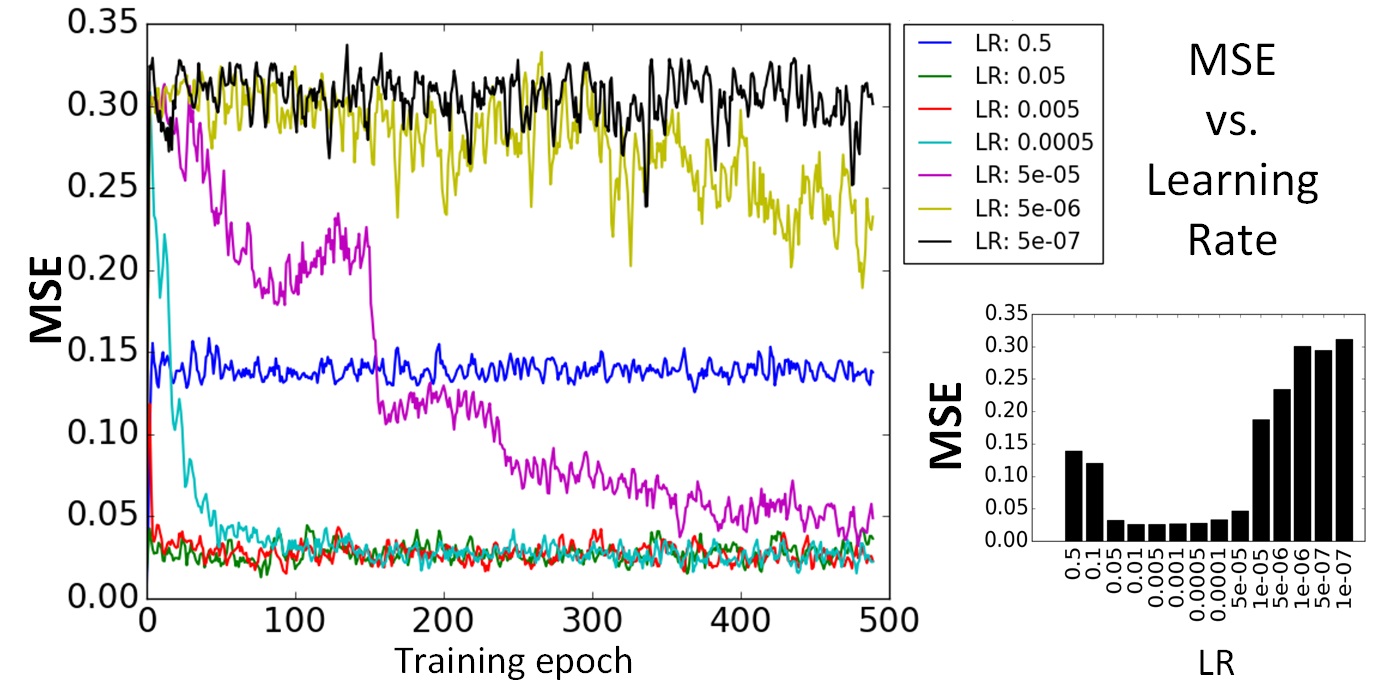}
\caption{MSE of the XOR learning process (left: through 500 iterations, right: after training).}
\label{fig:xorlr}
\end{figure}

Figure~\ref{fig:xorspike} shows the training process where each box represents the two output neurons' activities with respect to the four input spike patterns determining $\{0,1,1,0\}$ classes. After around 150 training iterations, the output neurons become selective to the input categories. Figure~\ref{fig:xorlr} demonstrates the learning convergence progress using the energy function defined in Eq.~\ref{eq:xorenergy}. This figure shows that the proper learning rates, $\mu$, fall in the range [0.01, 0.0005]. 
\begin{equation}
\label{eq:xorenergy}
\mathrm{MSE} = \frac{1}{N} \sum_{k=1}^N \big (\frac{1}{T}\sum_{t=1}^T \xi^k(t) \big )^2 \ , \ \ \xi^k(t) = \sum_{i} \xi^k_i(t)
\end{equation}
In Eq.~\ref{eq:xorenergy}, $N$ and $\xi^k_i(t)$ denote the training batch size and the error value of output neuron $i$ in response to sample $k$.

\subsection{Iris dataset}
The Iris dataset consists of three different types of flowers (Setosa, Versicolour, and Virginica) represented by the length and width of both petal and sepal (4 features). After normalizing the feature values in the range [0,1], input spike trains are generated. The network architecture in this experiment consists of 4 input, 30 hidden, and 3 output IF neurons. We found, in our preliminary experiments, that using more than 10 hidden neurons does not significantly improve accuracy. 

\begin{figure}
\centering
\includegraphics[scale=.5]{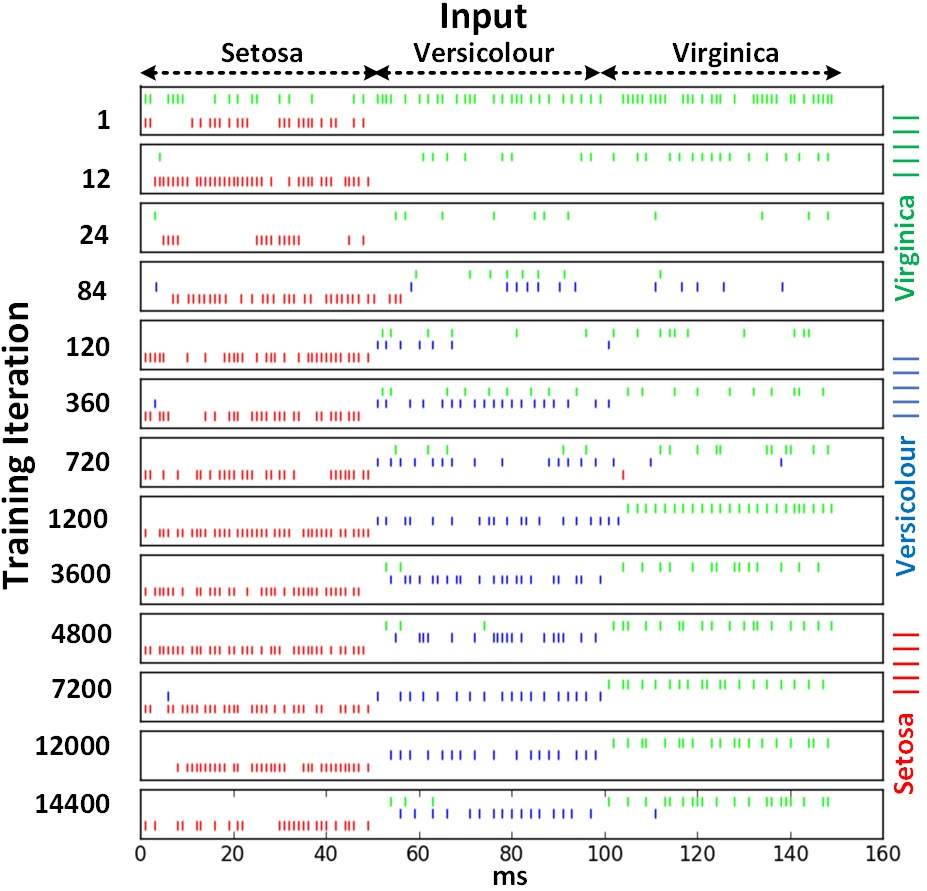}
\caption{Spike trains released from the output neurons in response to the spike trains representing Setosa, Versicolour, and Virginica.}
\label{fig:irisspike}
\end{figure} 

\begin{figure}
\centering
\includegraphics[scale=.42]{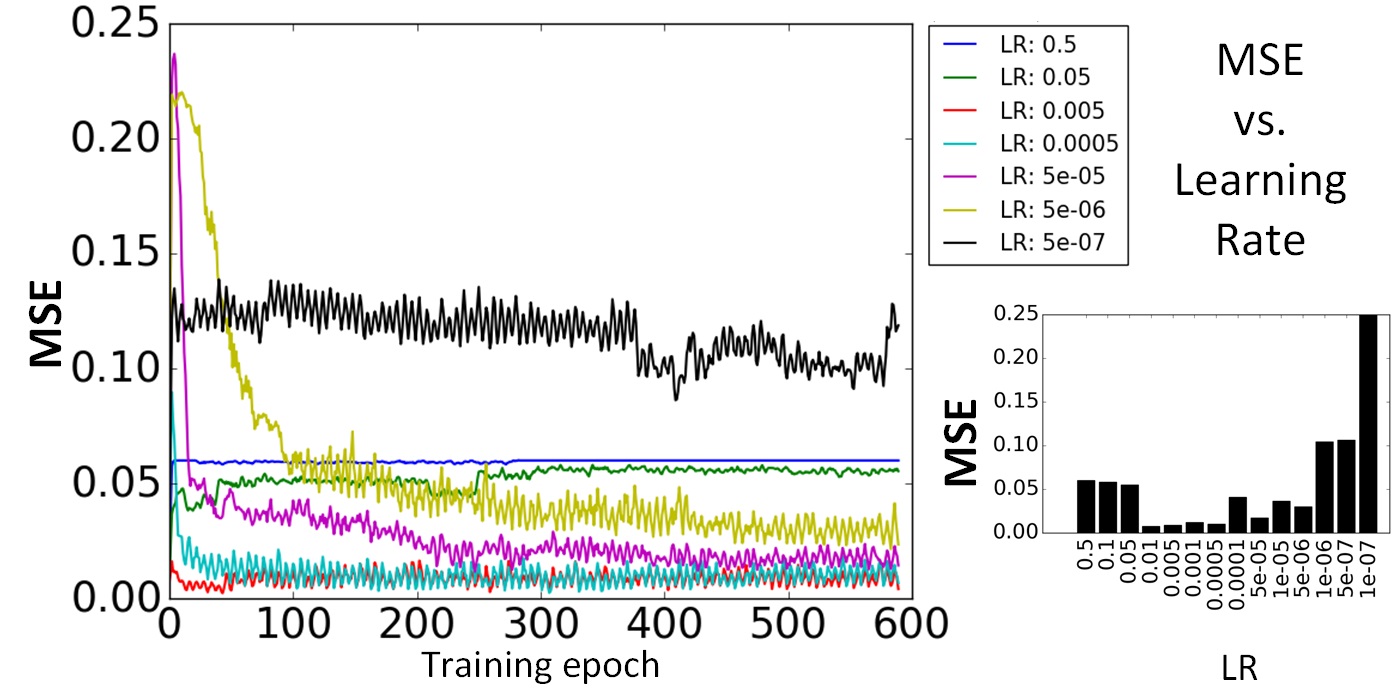}
\caption{MSE of the learning process applied to the Iris dataset (left: through 500 iterations, right: after training).}
\label{fig:irisenergy}
\end{figure} 

Similar to the results obtained for the XOR problem, Figure~\ref{fig:irisspike} illustrates that the SNN converges to be selective to the three flower patterns through training. Figure~\ref{fig:irisenergy} shows the learning process of the SNN in terms of the MSE defined in Eq.~\ref{eq:xorenergy}. The final evaluation using 5-fold cross validation reported 96\% accuracy while the traditional neural network equipped with standard Backpropagation showed 96.7\% accuracy. This result shows the success of the proposed BP-STDP algorithm to train temporal SNNs. Furthermore, Table~\ref{tb:iris} compares our results with other spike-based and traditional supervised learning methods. Our model outperforms (or equally performs) the previous multi-spiking supervised learning algorithms except Lin et al.'s method~\cite{lin2017supervised} where develops a spatio-temporal, computationally expensive GD. 

\begin{table}[]
\centering
\caption{Accuracy of Iris classification using our method (BP-STDP) in comparison with the other spiking supervised approaches and the classical ANN, SVM, and Naive Bayes methods.}
\label{tb:iris}
\begin{tabular}{|l|l|}
\hline
\textit{\textbf{Model}} & \textit{\textbf{Accuracy \%}} \\ \hline
SpikeProp (Bohte et al. 2002)~\cite{bohte2002error}                  & 96.1                          \\ \hline
MSGD; Xu et al. (2013)~\cite{xu2013supervised}                       & 94.4                         \\ \hline
Wang et al. (2014)~\cite{wang2014online}                      & 86.1                          \\ \hline
Yu et al. (2014)~\cite{yu2014brain}                      & 92.6                          \\ \hline
SWAT; Wade et al. (2010)~\cite{wade2010swat}                    & 95.3                          \\ \hline
multi-ReSuMe; Sporea et al. (2013)~\cite{sporea2013supervised}             & 94.0                          \\ \hline
Xie et al. (2016)~\cite{xie2016efficient}                       & 96.0                          \\ \hline
Lin et al. (2017)~\cite{lin2017supervised}                       & 96.7                          \\ \hline
SVM and Bayesian~\cite{wang2014online}       & 96.0                          \\ \hline
ANN (30 hidden neurons)                  & 96.7                          \\
\hline
\textbf{BP-STDP}        & \textbf{96.0}                 \\ \hline
\end{tabular}
\end{table}
 
\subsection{MNIST Dataset}
To assess the proposed algorithm in solving more complex problems, we evaluate the SNN on MNIST with 784 input, 100 through 1500 hidden, and 10 output IF neurons equipped with BP-STDP. The SNN was trained and tested on 60k training and 10k testing samples. The input spike trains are generated by random lags with the spike rates proportional to the normalized pixel values in the range [0, 1].

\begin{figure}
\centering
\includegraphics[scale=.34]{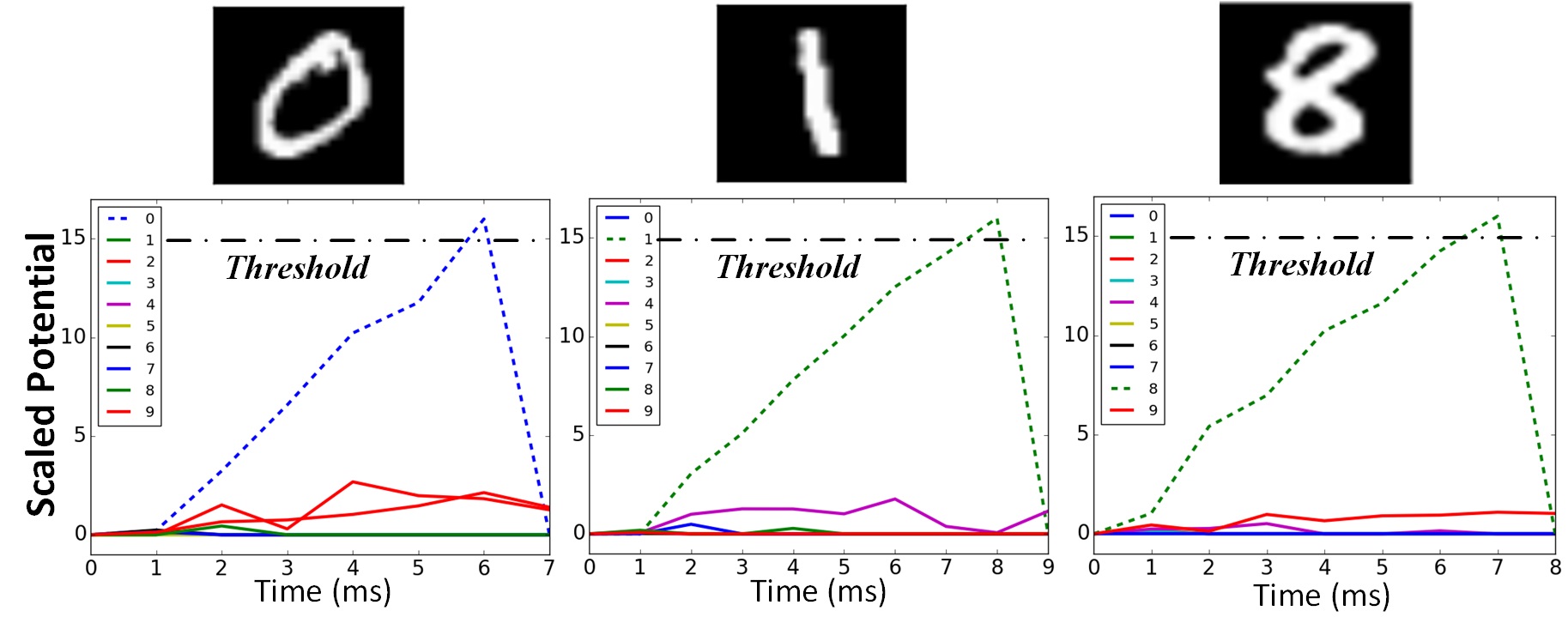}
\caption{Temporal membrane potentials of output neurons in response to randomly selected `0', `1', and `8' handwritten digits.}
\label{fig:mnistU}
\end{figure}

\begin{figure}
\centering
\subfloat[]
{
\includegraphics[scale=.365]{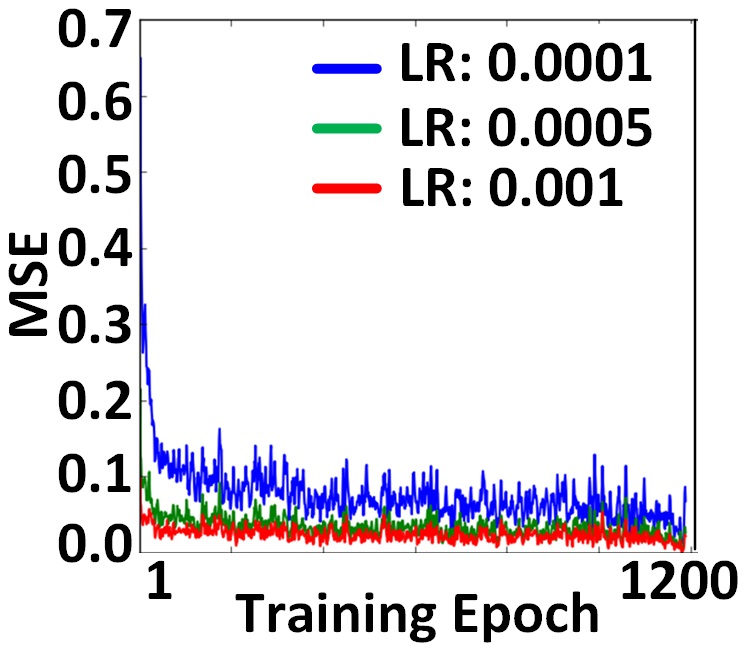}
\label{fig:mnistlr}
}
\quad
\subfloat[] {
\includegraphics[scale=.35]{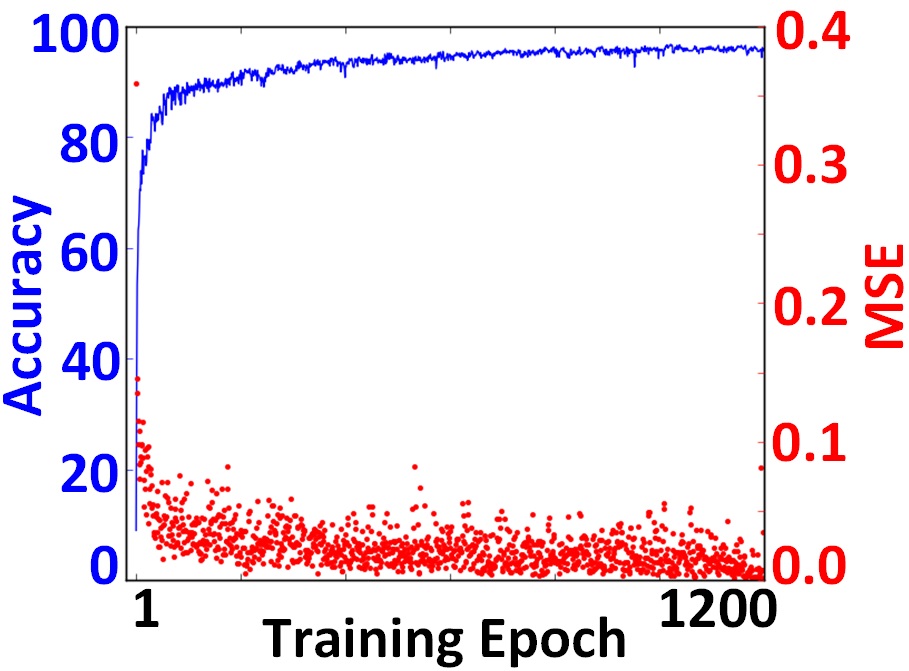}
\label{fig:mnistenergy}
}
\caption{(a): MSE of the SNN over training with $\mu=0.001, 0.0005, 0.0001$. (b): MSE and accuracy of the SNN with $\mu=0.0005$. $H=1000$ and each training epoch stands for 50 images.}
\end{figure}

\begin{figure}
\centering
\includegraphics[scale=.57]{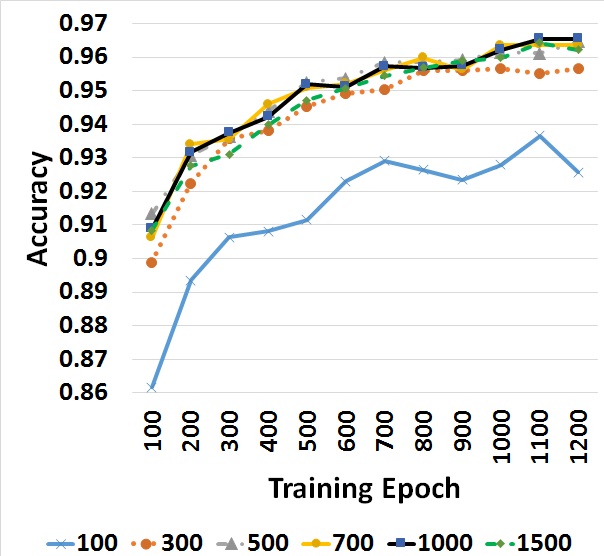}
\caption{Performance of BP-STDP applied to the SNN with 100 through 1500 hidden IF neurons over 100 through 1200 training epochs.}
\label{fig:mnisthidden}
\end{figure}

Figure~\ref{fig:mnistU} shows the output neurons' membrane potentials (after training) in response to the spike trains generated from three randomly selected digits. The target neuron's membrane potential grows fast and reaches the threshold while the other neurons' activities are near zero. Furthermore, this fast response ($<9$ ms) reduces the network's response latency.
Figure~\ref{fig:mnistlr} shows the learning process for 1200 training epochs. Each epoch stands for 50 MNIST digits. The MSE track in this plot shows the fast convergence of BP-STDP for the learning rates of 0.001 and 0.0005. Figure~\ref{fig:mnistenergy} shows the MSE values and the accuracy rates for the SNN with 1000 hidden neurons over training. After 100 and 900 training epochs, the performances of 90\% and 96\% are achieved. To examine the impact of the number of hidden neurons on performance, we applied the BP-STDP rule to six SNNs with 100 through 1500 hidden IF neurons. Figure~\ref{fig:mnisthidden} shows the accuracy rates over training for these SNNs. The best accuracy rates belong to the networks with more than 500 hidden neurons.  

Finally, the BP-STDP algorithm was evaluated on 2-layer and 3-layer SNNs. The SNN architectures (the number of hidden neurons) are the same as the neural network architectures utilized in~\cite{lecun1998gradient} to show a better comparison. Our model achieved $96.6\pm 0.1$\% and $97.2\pm 0.07$\% accuracy rates for the 2-layer and 3-layer SNNs, respectively. These results are comparable to the accuracy rates reported by traditional neural networks (trained by backprpagation)~\cite{lecun1998gradient}.   
Table~\ref{tab:mnist} compares the proposed supervised learning method (BP-STDP) with the traditional backpropagation algorithm (or GD), spiking backpropagation methods embedded in multi-layer (deep) SNNs, and recent STDP-based SNNs for MNIST classification. This comparison confirms the success of the bio-inspired BP-STDP rule applied to the temporal SNN architecture. 
Our model is the only high performance, end-to-end STDP based, supervised learning approach applied to multi layer SNNs. The SNNs introduced in~\cite{tavanaei2016bio,kheradpisheh2017stdp} develop multi-layer STDP learning for feature extraction. However, they use a support vector machine (SVM) classifier for the final supervised layer. Although the gradient descent approaches performed successfully~\cite{o2016deep,diehl2015unsupervised,lee2016training}, they do not offer bio-inspired, power-efficient STDP learning in SNNs.
Specifically, the recently developed deep SNN using backpropagation~\cite{lee2016training} showed higher accuracy than the BP-STDP's accuracy; however, it computes the derivatives of activation functions obtained by the neurons' membrane potentials instead of using their spike events. 

\begin{table}[]
\centering
\footnotesize
\caption{MNIST classification performance of the proposed BP-STDP applied to SNNs in comparison with 1) traditional backpropagation applied to conventional neural networks, 2) spiking gradient descent, and 3) recent STDP-based SNNs. `*' denotes a random distortion of training set (data augmentation for improving the performance). H1 and H2 show the number of neurons in the first and second hidden layers.}
\label{tab:mnist}
\begin{tabular}{|l|l|l|}
\hline
\multicolumn{1}{|c|}{\textbf{Model}} & \multicolumn{1}{c|}{\textbf{Description}}                                                                             & \multicolumn{1}{c|}{\textbf{Acc. (\%)}}                   \\ \hline
\multicolumn{3}{|c|}{\textit{Traditional Neural Networks}}                                                                                                                                                          \\ \hline
Lecun et al.~\cite{lecun1998gradient}                         & \begin{tabular}[c]{@{}l@{}}Traditional backpropagation.\\ 2-layer network; H1=300\end{tabular}                        & \begin{tabular}[c]{@{}l@{}}95.3\\ 96.4*\end{tabular} \\ \hline
Lecun et al.~\cite{lecun1998gradient}                         & \begin{tabular}[c]{@{}l@{}}Traditional backpropagation.\\ 2-layer network; H1=1000\end{tabular}                       & \begin{tabular}[c]{@{}l@{}}95.5\\ 96.2*\end{tabular} \\ \hline
Lecun et al.~\cite{lecun1998gradient}                         & \begin{tabular}[c]{@{}l@{}}Traditional backpropagation.\\ 3-layer network; H1=500; H2=150\end{tabular}                & \begin{tabular}[c]{@{}l@{}}97.1\\ 97.6*\end{tabular} \\ \hline
\multicolumn{3}{|c|}{\textit{Spiking Gradient Descent}}                                                                                                                                                             \\ \hline
O'Connor et al.~\cite{o2016deep}                      & \begin{tabular}[c]{@{}l@{}}Deep SNN.\\ Stochastic gradient descent\end{tabular}                                       & 96.40                                                \\ \hline
O'Connor et al.~\cite{o2016deep}                      & \begin{tabular}[c]{@{}l@{}}Deep SNN.\\ Fractional stochastic gradient descent\end{tabular}                            & 97.93                                                \\ \hline
Lee et al.~\cite{lee2016training}                           & \begin{tabular}[c]{@{}l@{}}Deep SNN. Backpropagation\\ Membrane potential as activation value\end{tabular}         & 98.88                                                \\ \hline
\multicolumn{3}{|c|}{\textit{STDP Based Models}}                                                                                                                                                               \\ \hline
Diehl et al.~\cite{diehl2015unsupervised}                         & \begin{tabular}[c]{@{}l@{}}2-Layer SNN.\\ STDP; Example-based classifier\end{tabular}                                 & 95.00                                                \\ \hline
Tavanaei et al.~\cite{tavanaei2016bio}                      & \begin{tabular}[c]{@{}l@{}}Spiking CNN.\\ Sparse coding and STDP; SVM classifier\end{tabular}                         & 98.36                                                \\ \hline
Kheradpisheh et al.~\cite{kheradpisheh2017stdp}                  & \begin{tabular}[c]{@{}l@{}}Spiking CNN.\\ Layer-wise STDP; SVM classifier\end{tabular}                                & 98.40                                                \\ \hline
Neftci et al.~\cite{neftci2014event}                        & \begin{tabular}[c]{@{}l@{}}Spiking restricted Boltzmann machine.\\ Contrastive divergence for IF neurons\end{tabular} & 91.90                                                \\ \hline
BP-STDP (This paper)                 & \begin{tabular}[c]{@{}l@{}}STDP-based backpropagation.\\ 2-layer SNN; H1=300\end{tabular}                             & 95.70                                                \\ \hline
BP-STDP (This paper)                 & \begin{tabular}[c]{@{}l@{}}STDP-based backpropagation.\\ 2-layer SNN; H1=1000\end{tabular}                            & 96.60                                                \\ \hline
\textbf{BP-STDP (This paper)}        & \textbf{\begin{tabular}[c]{@{}l@{}}STDP-based backpropagation.\\ 3-layer SNN; H1=500; H2=150\end{tabular}}            & \textbf{97.20}                                       \\ \hline
\end{tabular}
\end{table}

\section{Discussion}
BP-STDP introduces a novel supervised learning for SNNs. It showed promising performances comparable to the state-of-the-art conventional gradient descent approaches. BP-STDP provides bio-inspired local learning rules which take spike times into consideration as well as spike rates in the subsequent layers of IF neurons. Bengio et al.~\cite{bengio2017stdp} showed that the synaptic weight change is proportional to the presynaptic spike event and the postsynaptic temporal activity that is analogous to the STDP rule and confirms Hinton's idea that says STDP can be associated with the postsynaptic temporal rate~\cite{hinton2007backpropagation}. In this paper, we showed that the backpropagation update rules can be used to develop spatio-temporally local learning rules, which implement biologically plausible STDP in SNNs. 

The proposed algorithm is inspired from the backpropagation update rules used for the conventional networks of ReLU neurons. However, it develops biologically plausible, temporally local learning rules in an SNN. This matter was accomplished by an initial approximation of the IF neurons to the ReLU neurons to support the spike-based communication scheme in SNNs. The spiking supervised learning rules offer a combination of STDP and anti-STDP applied to spiking neural layers corresponding to the temporal presynaptic and postsynaptic neural activities. Therefore, we take advantages of both accurate gradient descent and efficient, temporally local STDP in spiking frameworks. The main question is that how does error propagation corresponds to the spiking behavior of IF neurons in multi-layer network architectures? To answer this question, let us assume the error value as a signal that either stimulates ($\sum_i \xi_i >0$) or suppresses ($\sum_i \xi_i <0$) the IF neuron to fire. Stimulating (suppressing) a neuron refers to increasing (decreasing) its membrane potential that is proportionally controlled by its input weights and presynaptic spike events. Thus, as the BP-STDP update rules change the synaptic weights temporally based on the error signal and the presynaptic spike times, it manipulates the neural activities (action potentials) of hidden layers at each time step.

The experimental results showed the success of BP-STDP in implementing supervised learning embedded in multi-layer SNNs. The XOR problem proved the ability of BP-STDP to classify non-linearly separable samples represented by spike trains through $T$ ms time intervals. The complex problems of IRIS and MNIST classification demonstrated comparable performances (96.0\% and 97.2\% respectively) to the conventional backpropagation algorithm and the recent SNNs while BP-STDP offers an end-to-end STDP-based supervised learning for spiking pattern classification. Using the bio-inspired and temporally local STDP rules guides us one step closer to the efficient computations that occur in the brain. To the best of our knowledge, this approach is the first high performance, STDP-based supervised learning while avoiding computationally expensive gradient descent.

\section{Conclusion}
This paper showed that IF neurons approximate rectified linear units, if the neurons' activities are mapped to the spike rates. Hence, a network of spiking IF neurons can undergo backpropagation learning applied to conventional NNs. We proposed a temporally local learning rule (derived from the traditional backpropagation updates) incorporating the STDP and anti-STDP rules embedded in a multi-layer SNN of IF neurons. This model (BP-STDP) takes advantages of the bio-inspired, efficient STDP rule in spiking platforms and the power of GD for training multi-layer networks. Also, converting the GD-based weight change rules to the spike-based STDP rules is much easier and computationally inexpensive than developing a spiking GD rule. The experiments on the XOR problem showed that the proposed SNN can classify non-linearly separable patterns. Furthermore, the final evaluations on the Iris and MNIST datasets demonstrated high classification accuracies comparable to the state-of-the-art, multi-layer networks of traditional and spiking neurons.

The promising results of the BP-STDP model warrants our future investigation to develop a deep SNN equipped with BP-STDP and regularization modules. The deep SNN can be utilized for larger pattern recognition tasks while preserving efficient, brain-like computations.

\section*{References}
\bibliographystyle{elsarticle-num} 
\bibliography{amir}

\begin{thebibliography}{10}
\expandafter\ifx\csname url\endcsname\relax
  \def\url#1{\texttt{#1}}\fi
\expandafter\ifx\csname urlprefix\endcsname\relax\def\urlprefix{URL }\fi
\expandafter\ifx\csname href\endcsname\relax
  \def\href#1#2{#2} \def\path#1{#1}\fi

\bibitem{maass1997networks}
W.~Maass, Networks of spiking neurons: the third generation of neural network
  models, Neural networks 10~(9) (1997) 1659--1671.

\bibitem{ghosh2009spiking}
S.~Ghosh-Dastidar, H.~Adeli, Spiking neural networks, International journal of
  neural systems 19~(04) (2009) 295--308.

\bibitem{maass2015spike}
W.~Maass, To spike or not to spike: that is the question, Proceedings of the
  IEEE 103~(12) (2015) 2219--2224.

\bibitem{neil2016learning}
D.~Neil, M.~Pfeiffer, S.-C. Liu, Learning to be efficient: Algorithms for
  training low-latency, low-compute deep spiking neural networks, in:
  Proceedings of the 31st Annual ACM Symposium on Applied Computing, ACM, 2016,
  pp. 293--298.

\bibitem{kasabov2013dynamic}
N.~Kasabov, K.~Dhoble, N.~Nuntalid, G.~Indiveri, Dynamic evolving spiking
  neural networks for on-line spatio-and spectro-temporal pattern recognition,
  Neural Networks 41 (2013) 188--201.

\bibitem{bohte2002error}
S.~M. Bohte, J.~N. Kok, H.~La~Poutre, Error-backpropagation in temporally
  encoded networks of spiking neurons, Neurocomputing 48~(1) (2002) 17--37.

\bibitem{mckennoch2006fast}
S.~McKennoch, D.~Liu, L.~G. Bushnell, Fast modifications of the spikeprop
  algorithm, in: Neural Networks, 2006. IJCNN'06. International Joint
  Conference on, IEEE, 2006, pp. 3970--3977.

\bibitem{booij2005gradient}
O.~Booij, H.~tat Nguyen, A gradient descent rule for spiking neurons emitting
  multiple spikes, Information Processing Letters 95~(6) (2005) 552--558.

\bibitem{ghosh2009new}
S.~Ghosh-Dastidar, H.~Adeli, A new supervised learning algorithm for multiple
  spiking neural networks with application in epilepsy and seizure detection,
  Neural networks 22~(10) (2009) 1419--1431.

\bibitem{xu2013supervised}
Y.~Xu, X.~Zeng, L.~Han, J.~Yang, A supervised multi-spike learning algorithm
  based on gradient descent for spiking neural networks, Neural Networks 43
  (2013) 99--113.

\bibitem{xu2017online}
Y.~Xu, J.~Yang, S.~Zhong, An online supervised learning method based on
  gradient descent for spiking neurons, Neural Networks 93 (2017) 7--20.

\bibitem{florian2012chronotron}
R.~V. Florian, The chronotron: a neuron that learns to fire temporally precise
  spike patterns, PloS one 7~(8) (2012) e40233.

\bibitem{victor1997metric}
J.~D. Victor, K.~P. Purpura, Metric-space analysis of spike trains: theory,
  algorithms and application, Network: computation in neural systems 8~(2)
  (1997) 127--164.

\bibitem{huh2017gradient}
D.~Huh, T.~J. Sejnowski, Gradient descent for spiking neural networks, arXiv
  preprint arXiv:1706.04698 (2017) 99.

\bibitem{van2001novel}
M.~C. van Rossum, A novel spike distance, Neural computation 13~(4) (2001)
  751--763.

\bibitem{zenke2017superspike}
F.~Zenke, S.~Ganguli, {SuperSpike}: Supervised learning in multi-layer spiking
  neural networks, arXiv preprint arXiv:1705.11146 (2017) 99.

\bibitem{lin2017supervised}
X.~Lin, X.~Wang, Z.~Hao, Supervised learning in multilayer spiking neural
  networks with inner products of spike trains, Neurocomputing 237 (2017)
  59--70.

\bibitem{lee2016training}
J.~H. Lee, T.~Delbruck, M.~Pfeiffer, Training deep spiking neural networks
  using backpropagation, Frontiers in neuroscience 10 (2016) 1--13.

\bibitem{wu2017spatio}
Y.~Wu, L.~Deng, G.~Li, J.~Zhu, L.~Shi, Spatio-temporal backpropagation for
  training high-performance spiking neural networks, arXiv preprint
  arXiv:1706.02609 (2017) 99.

\bibitem{xie2016efficient}
X.~Xie, H.~Qu, G.~Liu, M.~Zhang, J.~Kurths, An efficient supervised training
  algorithm for multilayer spiking neural networks, PloS one 11~(4) (2016)
  e0150329.

\bibitem{ponulak2010supervised}
F.~Ponulak, A.~Kasi{\'n}ski, Supervised learning in spiking neural networks
  with {ReSuMe}: sequence learning, classification, and spike shifting, Neural
  Computation 22~(2) (2010) 467--510.

\bibitem{sporea2013supervised}
I.~Sporea, A.~Gr{\"u}ning, Supervised learning in multilayer spiking neural
  networks, Neural computation 25~(2) (2013) 473--509.

\bibitem{mohemmed2012span}
A.~Mohemmed, S.~Schliebs, S.~Matsuda, N.~Kasabov, {SPAN}: Spike pattern
  association neuron for learning spatio-temporal spike patterns, International
  Journal of Neural Systems 22~(04) (2012) 1250012.

\bibitem{mohemmed2013training}
A.~Mohemmed, S.~Schliebs, S.~Matsuda, N.~Kasabov, Training spiking neural
  networks to associate spatio-temporal input--output spike patterns,
  Neurocomputing 107 (2013) 3--10.

\bibitem{gutig2006tempotron}
R.~G{\"u}tig, H.~Sompolinsky, The tempotron: a neuron that learns spike
  timing-based decisions, Nature neuroscience 9~(3) (2006) 420.

\bibitem{yu2014brain}
Q.~Yu, H.~Tang, K.~C. Tan, H.~Yu, A brain-inspired spiking neural network model
  with temporal encoding and learning, Neurocomputing 138 (2014) 3--13.

\bibitem{xu2013new}
Y.~Xu, X.~Zeng, S.~Zhong, A new supervised learning algorithm for spiking
  neurons, Neural computation 25~(6) (2013) 1472--1511.

\bibitem{markram2012spike}
H.~Markram, W.~Gerstner, P.~J. Sj{\"o}str{\"o}m, Spike-timing-dependent
  plasticity: a comprehensive overview, Frontiers in synaptic neuroscience 4
  (2012) 1--3.

\bibitem{song2000competitive}
S.~Song, K.~D. Miller, L.~F. Abbott, Competitive hebbian learning through
  spike-timing-dependent synaptic plasticity, Nature neuroscience 3~(9) (2000)
  919--926.

\bibitem{caporale2008spike}
N.~Caporale, Y.~Dan, Spike timing--dependent plasticity: a hebbian learning
  rule, Annu. Rev. Neurosci. 31 (2008) 25--46.

\bibitem{wang2014online}
J.~Wang, A.~Belatreche, L.~Maguire, T.~M. Mcginnity, An online supervised
  learning method for spiking neural networks with adaptive structure,
  Neurocomputing 144 (2014) 526--536.

\bibitem{tavanaei2017spiking}
A.~Tavanaei, A.~S. Maida, A spiking network that learns to extract spike
  signatures from speech signals, Neurocomputing 240 (2017) 191--199.

\bibitem{bishop2006pattern}
C.~M. Bishop, Pattern recognition and machine learning, springer, 2006.

\bibitem{goodfellow2016deep}
I.~Goodfellow, Y.~Bengio, A.~Courville, Deep learning, MIT press, 2016.

\bibitem{tavanaei2017representation}
A.~Tavanaei, T.~Masquelier, A.~Maida, Representation learning using event-based
  {STDP}, arXiv preprint arXiv:1706.06699 (2017) 99.

\bibitem{fisher1936use}
R.~A. Fisher, The use of multiple measurements in taxonomic problems, Annals of
  human genetics 7~(2) (1936) 179--188.

\bibitem{lecun1998mnist}
Y.~LeCun, The {MNIST} database of handwritten digits, http://yann. lecun.
  com/exdb/mnist/.

\bibitem{wade2010swat}
J.~J. Wade, L.~J. McDaid, J.~A. Santos, H.~M. Sayers, Swat: a spiking neural
  network training algorithm for classification problems, IEEE Transactions on
  Neural Networks 21~(11) (2010) 1817--1830.

\bibitem{lecun1998gradient}
Y.~LeCun, L.~Bottou, Y.~Bengio, P.~Haffner, Gradient-based learning applied to
  document recognition, Proceedings of the IEEE 86~(11) (1998) 2278--2324.

\bibitem{tavanaei2016bio}
A.~Tavanaei, A.~S. Maida, Bio-inspired spiking convolutional neural network
  using layer-wise sparse coding and {STDP} learning, arXiv preprint
  arXiv:1611.03000 (2016) 99.

\bibitem{kheradpisheh2017stdp}
S.~R. Kheradpisheh, M.~Ganjtabesh, S.~J. Thorpe, T.~Masquelier, {STDP}-based
  spiking deep convolutional neural networks for object recognition, Neural
  Networks (2017) 56--67.

\bibitem{o2016deep}
P.~O'Connor, M.~Welling, Deep spiking networks, arXiv preprint arXiv:1602.08323
  (2016) 99.

\bibitem{diehl2015unsupervised}
P.~U. Diehl, M.~Cook, Unsupervised learning of digit recognition using
  spike-timing-dependent plasticity, Frontiers in computational neuroscience 9
  (2015) 99.

\bibitem{neftci2014event}
E.~Neftci, S.~Das, B.~Pedroni, K.~Kreutz-Delgado, G.~Cauwenberghs, Event-driven
  contrastive divergence for spiking neuromorphic systems, Frontiers in
  neuroscience 7 (2014) 272.

\bibitem{bengio2017stdp}
Y.~Bengio, T.~Mesnard, A.~Fischer, S.~Zhang, Y.~Wu, {STDP}-compatible
  approximation of backpropagation in an energy-based model, Neural computation
  (2017) 555--577.

\bibitem{hinton2007backpropagation}
G.~Hinton, How to do backpropagation in a brain, in: Invited talk at the
  NIPS’2007 Deep Learning Workshop, Vol. 656, 2007.

\end{thebibliography}


\end{document}